\documentclass[letterpaper, 10 pt, conference]{ieeeconf}  

\IEEEoverridecommandlockouts                              

\usepackage{amsmath} 
\usepackage{amssymb}  

\usepackage{graphicx}
\usepackage[adjust]{cite}
\usepackage[hidelinks]{hyperref}
\usepackage{url}
\usepackage{xspace}
\usepackage{siunitx}
\usepackage{tikz}
\usepackage{xcolor, colortbl}
\definecolor{Gray}{gray}{0.9}
\usepackage{booktabs}

\title{\LARGE \bf
Automatically designing robot swarms in environments populated by other robots: an experiment in robot shepherding
}

\author{David Garzón Ramos and Mauro Birattari
\thanks{The authors are with IRIDIA, Université libre de Bruxelles, Belgium. 
{\tt\small \{david.garzon.ramos, mauro.birattari\}@ulb.be}.} %
\thanks{The project has received funding from the European Research Council (ERC) under the European Union’s Horizon 2020 research and innovation programme (DEMIURGE: grant No.~681872) and from Belgium’s Wallonia-Brussels Federation through a ARC Advanced Project 2020 (Guaranteed by Optimization). The authors acknowledge support from Toyota Motor Europe. DGR acknowledges support from Minciencias, Colombia; MB from the Belgian Fonds de la Recherche Scientifique-FNRS.} 
\thanks{The authors thank the designers who participated in this study. Implementations and experiments were done by DGR. The paper was drafted by DGR and refined by DGR and MB. The research was directed by MB.}%
}

\newcommand{\etal}[1]{#1\,{\emph{et al.}}\xspace}

\newcommand{\argos}{ARGoS3\xspace}
\newcommand{\auto}{Au\-to\-MoDe\xspace}

\newcommand{\tutti}{\texttt{Tu\-tti\-Fru\-tti}\xspace}

\newcommand{\pist}{\texttt{Pis\-ta\-cchio}\xspace}

\newcommand{\evcol}{\texttt{EvoColor}\xspace}

\newcommand{\evcmy}{\texttt{EvoCMY}\xspace}

\newcommand{\cmaes}{\texttt{CMA-ES}\xspace}
\newcommand{\xnes}{\texttt{xNES}\xspace}
\newcommand{\neat}{\texttt{NEAT}\xspace}

\newcommand{\rwalk}{\texttt{R-Walk}\xspace}
\newcommand{\chuma}{\texttt{C-Human}\xspace}

\newcommand{\epuck}{e-puck\xspace}
\newcommand{\epucks}{e-pucks\xspace}

\newcommand{\irace}{Iterated F-race\xspace}

\newcommand{\Magg}{\textsc{Aggre\-gation}\xspace}
\newcommand{\Mdis}{\textsc{Dis\-persion}\xspace}
\newcommand{\Mher}{\textsc{Herding}\xspace}

\newcommand{\Batra}{\textsl{C1-Attrac\-tion}\xspace}
\newcommand{\Brepu}{\textsl{C2-Re\-pul\-sion}\xspace}
\newcommand{\Batre}{\textsl{C3-Attrac\-tion\&Re\-pul\-sion}\xspace}

\newcommand{\RMD}[1]{RM\,#1.3\xspace}

\newcommand{\modSto}{\textsc{stop}\xspace}

\newcommand{\modExp}{\textsc{exploration}\xspace}
\newcommand{\modElu}{\textsc{color-elusion}\xspace}
\newcommand{\modFol}{\textsc{color-following}\xspace}
\newcommand{\modCir}{\textsc{circling}\xspace}

\newcommand{\modFpr}{\textsc{fixed-probability}\xspace}
\newcommand{\modCol}{\textsc{color-detection}\xspace}
\newcommand{\modBla}{\textsc{black-floor}\xspace}
\newcommand{\modGra}{\textsc{gray-floor}\xspace}
\newcommand{\modWhi}{\textsc{white-floor}\xspace}

\begin{document}

\maketitle
\thispagestyle{empty}
\pagestyle{empty}

\begin{abstract}

Automatic design is a promising approach to realizing robot swarms.
Given a mission to be performed by the swarm, an automatic method produces the required control software for the individual robots.
Automatic design has concentrated on missions that a swarm can execute independently, interacting only with a static environment and without the involvement of other active entities.
In this paper, we investigate the design of robot swarms that perform their mission by interacting with other robots that populate their environment.
We frame our research within robot shepherding:
the problem of using a small group of robots---the shepherds---to coordinate a relatively larger group---the sheep.
In our study, the group of shepherds is the swarm that is automatically designed, and the sheep are pre-programmed robots that populate its environment.
We use automatic modular design and neuroevolution to produce the control software for the swarm of shepherds to coordinate the sheep.
We show that automatic design can leverage mission-specific interaction strategies to enable an effective coordination between the two groups.

\end{abstract}

\section{INTRODUCTION}\label{sec:introduction}

Swarm robotics~\cite{Ham2018book} is an approach to the coordination of large groups of robots~\cite{Sah2005sab}.
In a robot swarm~\cite{Ben2005sab,Sah2005sab}, a collective behavior emerges from the interaction among robots, and between the robots and their environment.
The design of collective behaviors for robot swarms is challenging:
the mission to be performed is defined at a global level, but the robots must be programmed at the individual level~\cite{BraFerBirDor2013SI}.
No generally applicable methodology exists to tell what an individual should do so that a collective behavior emerges in the swarm.
Typically, designers of robot swarms rely on manual trial-and-error processes to produce the control software of the robots~\cite{StoVarSvoBel2020FRAI}.
Alternatively, this problem has been addressed from the perspective of automatic design~\cite{BirLigBoz-etal2019FRAI,BirLigHas2020NATUMINT,FraBir2016FRAI}:
an optimization process designs the collective behavior of the robots by maximizing a mission-specific performance metric.
In this paper, we study the automatic design of robot swarms that operate in environments populated by other robots.
We frame this problem into the robot shepherding problem~\cite{LieBaySow-etal2004icra}.
In robot shepherding, it is assumed that two groups of robots of different kind operate in the same environment---the shepherds and the sheep.
Shepherds and sheep influence each other's behavior and constitute a.
heterogeneous system that must perform missions collectively.
We use automatic design methods to produce the control software of the shepherds so that they coordinate the sheep in a set of spatially-organizing missions~\cite{BraFerBirDor2013SI,SchUmlSenElm2020FRAI}.
The sheep operate with pre-defined fixed control software.
In a sense, the sheep are reactive robots that populate the environment of the automatically designed swarm of shepherds.

Our goal is to investigate whether an automatic design process can effectively identify and exploit the dynamics between the two groups of robots.
In our experiments, the behavior of the sheep is a black-box to the design process.
The dynamics of the sheep must be discovered while the automatic design process produces the control software of the shepherds.
To investigate this problem, we implement two automatic design methods, \pist and \evcmy, and we use them to design the control software of shepherds.
We test these two methods in nine experimental scenarios that combine shepherding missions with sheep that operate with diverse predefined control software.

We report qualitative and quantitative results of experiments conducted in simulation.
Along with the results obtained with \pist and \evcmy, we also report baseline results obtained with a manual design approach and with a simple random walk.
The results show that the automatic design methods are effective in identifying and exploiting the dynamics of the scenario to perform the missions at hand.
The control software automatically designed performs better than the baselines.

\section{RELATED WORK}\label{sec:soa}

The traditional approach to the automatic design of robot swarms is neuroevolution~\cite{NolFlo2000book,Tri2008book,Nol2021book}.
In this approach, an evolutionary process fine-tunes an artificial neural network that serves as control software for the robots.
In the last decade, automatic modular design (\auto)~\cite{FraBraBru-etal2014SI,BirLigFra2021admlsa} has been proposed as an alternative to neuroevolution.
In the \auto approach, an optimization process fine-tunes and selects software modules that are assembled into a predefined control architecture.
Researchers have investigated various aspects of the automatic design process with the two approaches: 
design strategies~\cite{ChrDor2006alife,JonWinHauStu2019AIS,GhaKucGarBir2023icra-up}, viable optimization algorithms~\cite{FraBraBru-etal2015SI,CamFer2022gecco}, control architectures~\cite{FraBraBru-etal2014SI,KucVanBir2021evoapps}, robot platforms~\cite{DuaCosGom-etal2016PLOSONE,GarBir2020AS,SalGarBir2024COMMENG,KegGarHas-etal2024RAL}, emergent collective behaviors~\cite{TriGroLab-etal2003ecal,FerTurDue-etal2015PLOSCB,MenGarMor-etal2022SWEVO}, among others.
Most swarm robotics research is still conducted under laboratory conditions: mainly, in controlled, static environments and with homogeneous systems~\cite{HamSchElm-etal2020FRAI}. 
These scenarios differ from what one could expect in future real deployments of robot swarms~\cite{DorTheTri2020SCIROB,DorTheTri2021PIEEE}---with highly dynamical environments that are populated by robots and other agents~\cite{KinPorStr-etal2023}.
So far, little research has been devoted to investigating how well automatic design can tackle missions that occur in such populated environments.

Robot shepherding has been investigated mostly within the framework of collective motion~\cite{GenSto2014ants,GenSto2016aamas,OzdGauGro2017ecal,LicBelDix2019IEEETR,PieSch2018IEEETR,HuTurKra-etal2020IEEETCDS,SebMonSag2022IEEETR}.
In most cases, the behavior of the sheep is inspired by \emph{flocking} behaviors.
Typically, a designer models the desired behavior for the sheep and the shepherd and manually produces control software for the two.
The designer uses their knowledge and expertise to tailor the behavior of the shepherds to the behavior of the sheep, which is assumed to be known at design time.
Typically, the goal of these studies is to verify whether the models are suitable for creating shepherding behaviors under specific conditions---like with models of other swarm behaviors in the literature~\cite{BraFerBirDor2013SI,NedSil2019SWEVO,SchUmlSenElm2020FRAI,DorTheTri2021PIEEE}.

Recent studies have demonstrated that creating such models is a viable principled method to produce specific shepherding behaviors. 
These studies show that it is possible to coordinate the sheep with a single shepherd~\cite{GenSto2014ants,GenSto2016aamas,LicBelDix2019IEEETR} or with a group of shepherds that act cooperatively~\cite{OzdGauGro2017ecal,PieSch2018IEEETR,HuTurKra-etal2020IEEETCDS,DosOzdGauGro2022ants,SebMonSag2022IEEETR}.
Some of these studies have shown the benefits of using optimization processes to fine-tune the parameters of the control software~\cite{OzdGauGro2017ecal,DosOzdGauGro2022ants}. 
However, these methods are not conceived to be of general applicability: they are unable to address different types of missions with varied interactions between shepherds and sheep.
Indeed, they cannot be directly transferred from one problem to another if the behavior of the sheep varies considerably.
For example, a model designed for sheep that move away from the shepherds cannot be ported to problems in which the sheep move towards them.

We show here that automatic design is a general framework for generating shepherding behaviors.
In this study, we address varied interactions between shepherds and sheep, without having to apply mission-specific modifications to the design method. 
We contend that this is a more generally-applicable approach to designing a robot swarm that must interact with other robots in its environment.

\section{AUTOMATIC DESIGN METHODS}\label{sec:methods}

We produce control software for the shepherds with two automatic design methods: \pist and \evcmy. See the supplementary material for an illustrative representation of the methods and the robot~\cite{GarBir2024icra-supp}.

\subsection{Robot platform}

\pist and \evcmy produce control software for an extended version of the \epuck robot~\cite{MonBonRae-etal2009arsc, GarFraBru-etal2015techrep}.
In this paper, we consider an \epuck whose capabilities are formally defined by the reference model \RMD{3}, see Table~\ref{tab:RM33}.
A \emph{reference model}, in this case \RMD{3}, defines the interface between the control software and the hardware of the robot.
The \epuck defined by \RMD{3} has eight proximity sensors, three ground sensors, an omni-directional vision system, two wheels, and three RGB LEDs.
The robot can perceive objects and other robots in a \SI{3}{cm} range with its proximity sensors $(\mathit{prox}_{i})$.
It can also detect whether the color of the floor is black, gray, or white with the ground sensors $(\mathit{gnd}_{j})$.
Using its vision system $(\mathit{cam}_{c})$, the robot can perceive peer robots that display the colors cyan, magenta, and yellow with their LEDs.
These peers can be perceived in a range of $\SI{0.40}{\meter}$ and with a $\ang{360}$ field of view.
For each color, there is a vector $(V_{c})$ that aggregates the relative position of all the robots perceived.
The control software of the robot can set the target velocity $(v_{k})$ of the two wheels between $\num{-0.12}$ and $\SI{0.12}{\meter\per\second}$.
The control software can also set the LEDs $(\mathit{LEDs})$ of the robot to display the colors cyan~$(\mathit{C})$, magenta~$(\mathit{M})$, and yellow~$(\mathit{Y})$, or none~$(\mathit{\varnothing})$.

We conceived \RMD{3} to limit the interactions between shepherds and sheep to those that are triggered by visually perceivable stimuli---i.e., the relative distance between robots and the colors they display. 
In this way, we simplify the visualization and monitoring of the interactions between the two groups of robots.

\begin{table}
    \vspace*{0.15cm}
	
	\caption{\label{tab:RM33}Reference model \RMD{3}.}
	\centering
	
	\begin{tabular}{>{\raggedright}p{1.85cm}>{\raggedright}p{2.55cm}>{\raggedright}p{3.0cm}}
		\toprule
		\textbf{Input} & \textbf{Value} & \textbf{Description} \tabularnewline
		\midrule
		$\mathit{prox}_{i\in\{1,\dots,8\}}$ & $[0,1]$ & proximity readings \tabularnewline
		$\mathit{gnd}_{j\in\{1,\dots,3\}}$ & $\{\mathit{black},\mathit{gray},\mathit{white}\}$ & ground readings \tabularnewline
		$\mathit{cam}_{c\in\{C,M,Y\}}$ & $\{\mathit{yes},\mathit{no}\}$ & color signal perceived \tabularnewline
		$V_{c\in\{C,M,Y\}}$ & $\langle\num{1.0}; (0,2\pi)\,\unit{\radian}\rangle$ & direction of color signal\tabularnewline
		\midrule
		\textbf{Output} & \textbf{Value} & \textbf{Description} \tabularnewline
		\midrule
		$v_{k\in\{l,r\}}$ &$\SIrange{-0.12}{0.12}{\meter\per\second}$ & target velocity \tabularnewline
		$\mathit{LEDs}$ & $\{\varnothing,C,M,Y\}$ & color signal emitted \tabularnewline
        \midrule
        \multicolumn{3}{l}{Period of the control cycle: $\SI{0.1}{\second}$} \tabularnewline
		\bottomrule 
	\end{tabular}
\end{table}

\pist and \evcmy produce control software that can be transferred between the simulated \epuck and its physical counterpart.
In this study, we have only considered the simulated version of the \epuck~\cite{GarFraBru-etal2015techrep}.
We conduct our experiments with \argos~\cite{PinTriOgr-etal2012SI}: a swarm robotics simulator that has been used in the past to automatically design collective behaviors that have been subsequently validated in real-robot experiments~\cite{FraBraBru-etal2014SI,TriLop2015PLOSONE,FerTurDue-etal2015PLOSCB,FraBraBru-etal2015SI,GarBir2020AS,HasLigRudBir2021NATUCOM}.

\subsection{\pist}
\pist is an automatic design method that belongs to the \auto family~\cite{FraBraBru-etal2014SI,BirLigFra2021admlsa}.
It fine-tunes, selects, and combines parametric software modules into probabilistic finite-state machines via an optimization process.
\pist's modules are adapted from those of \tutti~\cite{GarBir2020AS}
and enable the \epuck to interact with its peers by perceiving and displaying colors.
\pist integrates five low-level behavior modules---\modExp, \modSto, \modFol, \modElu, and \modCir---and five transition conditions---\modBla, \modGra, \modWhi, \modFpr, and \modCol.
The low-level behaviors are actions that the robot can execute, and the transition conditions are events that can trigger the transition between low-level behaviors.
Table~\ref{tab:modules} lists \pist's software modules and their parameters.

\begin{table}
    \vspace*{0.15cm}
	
	\caption{\label{tab:modules} \pist's software modules.}
	\centering
	\begin{tabular}{>{\raggedright}p{2.45cm}>{\raggedright}p{1.3cm}>{\raggedright}p{3.65cm}}
		\toprule
		\textbf{Behavior} & \textbf{Parameter} & \textbf{Description}\tabularnewline
		\midrule
		\modExp & $\{\tau,\gamma\}$ & random walk with $\tau$ rotations\tabularnewline
		\modSto & $\{\gamma\}$ & standstill state\tabularnewline
		\modFol & $\{\delta,\gamma\}$ & attraction to color $\delta$ \tabularnewline
		\modElu & $\{\delta,\gamma\}$ & repulsion from color $\delta$ \tabularnewline
		\modCir & $\{\theta,\gamma\}$ & circular movement with angle $\theta$ \tabularnewline
		\midrule
		\textbf{Transition}& \textbf{Parameter} & \textbf{Condition}\tabularnewline
		\midrule
		\modBla &$\{\beta\}$& detected black floor \tabularnewline
		\modGra &$\{\beta\}$& detected gray floor \tabularnewline
		\modWhi &$\{\beta\}$& detected white floor\tabularnewline
		\modFpr &$\{\beta\}$& fixed probability transition\tabularnewline
		\modCol &$\{\delta,\beta\}$& detected signal of color $\delta$ \tabularnewline
		\midrule
        \multicolumn{3}{l}{All low-level behaviors can also set the LEDs with $\gamma\in\{\varnothing,C,M,Y\}$} \tabularnewline
        \multicolumn{3}{l}{$\beta$ is the probability for a transition to occur if the condition is fulfilled} \tabularnewline
        \midrule
        \multicolumn{3}{l}{\pist automatically tunes the parameters $\tau$, $\theta$, $\delta$, $\gamma$, and $\beta$} \tabularnewline
        \bottomrule
	\end{tabular}
\end{table}

Given a mission-specific performance metric, \pist uses an optimization algorithm to search for configurations of the control software that maximize the performance of the swarm. 
The control software has the form of a probabilistic-finite state machine with a maximum of four states---the low-level behavior modules---and a maximum of four outgoing transitions per state---the transition modules.
A transition always originates and ends in different states.
The optimization algorithm used in \pist is \irace~\cite{LopDubPer-etal2016ORP}---a \emph{de facto} standard optimization algorithm in \auto methods.
\irace builds finite-states machines and assesses their performance through simulations in \argos.
The duration of the optimization process is restricted by a predefined simulation budget.
When the simulations budget is exhausted, the design process ends and \pist returns the best configuration of control software it found.
We test the produced control software in the shepherds without any modification.

\subsection{\evcmy}
\evcmy is a straightforward implementation of the neuroevolutionary approach~\cite{NolFlo2000book,Tri2008book,Nol2021book}.
It produces artificial neural networks whose synaptic weights are tuned via artificial evolution.
\evcmy is an adaptation of \evcol~\cite{GarBir2020AS}---a method to automatically design collective behaviors for robots that can display and perceive colors.
In this sense, like \pist, \evcmy produces control software for \epucks defined by \RMD{3}. 
Table~\ref{tab:evocmy} summarizes the neural network topology and artificial evolution parameters of \evcmy.
We do not consider more advanced neuroevolutionary implementations (e.g., \cmaes~\cite{HanOst2001EC}, \xnes~\cite{GlaSchYi-etal2010gecco}, and \neat~\cite{StaMii2002EC}) as research has shown that they do not provide performance advantages when applied off the shelf~\cite{HasLigRudBir2021NATUCOM,KegGarHas-etal2024RAL}.

\begin{table}
    \vspace*{0.15cm}
	\caption{\label{tab:evocmy} \evcmy's network topology and evolution parameters.}
	\centering
	\begin{tabular}{>{\raggedright}p{2.5cm}>{\raggedright}p{5.2cm}}
		\toprule
        \textbf{Architecture:}  \tabularnewline
        \multicolumn{2}{l}{Fully-connected feed-forward neural network without hidden layers} \tabularnewline        
        \midrule
		\textbf{Input nodes}& \textbf{Description} \tabularnewline
		\midrule
		$\mathit{in}_{a\in\{1,\dots,8\}}$ & proximity readings $\mathit{prox}_{i\in\{1,\dots,8\}}$ \tabularnewline
		$\mathit{in}_{a\in\{9,\dots,11\}}$ & ground readings $\mathit{gnd}_{j\in\{1,\dots,3\}}$ \tabularnewline
		$\mathit{in}_{a\in\{12,\dots,23\}}$ & scalar projections of $V_{c\in\{C,M,Y\}}$ \tabularnewline
		$\mathit{in}_{a\in\{24\}}$ & \emph{bias} node \tabularnewline
		\midrule
		\textbf{Output nodes} & \textbf{Mapping} \tabularnewline
		\midrule
		$\mathit{out}_{b\in\{1,\dots,4\}}$ & velocities $v_{k\in\{l,r\}}$ \tabularnewline
		$\mathit{out}_{b\in\{5,\dots,8\}}$ & color displayed by the LEDs, $\{\varnothing,C,M,Y\}$ \tabularnewline
		\midrule
		\textbf{Connections} & \textbf{Description} \tabularnewline
		\midrule
		$\mathit{conn}_{s\in\{1,\dots,192\}}$ & synaptic weights, with $\omega{\in}[-5,5]$  \tabularnewline
		\midrule
		Generations* & computed over the simulations budget\tabularnewline
		Population size & $100$\tabularnewline
		Elitism rate & $20$ \tabularnewline
		Mutation rate & $80$\tabularnewline
		\midrule
        \multicolumn{2}{l}{\evcmy automatically tunes the synaptic weights} \tabularnewline
        \bottomrule
	\end{tabular}
\end{table}

\evcmy uses artificial evolution to search for configurations of the neural network that generate good-performing behaviors.
The performance is measured with respect to a mission-specific performance metric that is given as part of the mission specification---the same as for \pist.
Also, like \irace in \pist, the evolutionary process in \evcmy uses simulations in \argos to assess the performance of candidate neural networks.
The evolutionary process applies elitism and mutation operators to fine-tune the synaptic weights so that the performance of the swarm is optimized.
The artificial evolution finishes when a simulations budget is exhausted.
Also in this case, we test the produced control software in the shepherds without any modification.

\section{EXPERIMENTAL SETUP}  \label{sec:experiments}
We considered an heterogeneous system of five shepherds and ten sheep, which jointly performed a set of missions.
We devised the control software for the sheep so that they did not take action unless stimulated by the shepherds.
In this way, the performance of the heterogeneous system strictly depended on the effectiveness of the shepherding behaviors that were designed.

\subsection{Sheep control software}
Each sheep operated with one out of three pre-defined instances of control software: \Batra, \Brepu, and \Batre.
Shepherds could stimulate the sheep by physical proximity in a \SI{3}{\cm} range, or by displaying colors with their LEDs in a \SI{40}{\cm} range.
The ground sensor of the sheep allowed them to detect regions of interest in the environment.

In \Batra, the sheep were attracted to shepherds that display the color magenta.
In \Brepu, the sheep were repelled from shepherds that display the color cyan.
In \Batre, the sheep were both attracted to shepherds that display the color magenta and repelled from shepherds that display the color cyan.
In all cases, the sheep displayed the color yellow and remained static if no stimuli were perceived---physical proximity or color.
If the sheep stepped into a white-floor region, it halted its movement and turned off its LEDs until the end of the mission.

\Batra, \Brepu, and \Batre are probabilistic finite-state machines created with software modules that are similar to those of \pist---see supplementary material.
We followed a manual trial-and-error process to design them.
These finite-state machines were undisclosed to the design process that generates the behavior of the shepherds, which sees them as a black box.

\subsection{Missions}

\begin{figure}
    \vspace*{0.15cm}
	\centering			
	\setlength{\fboxsep}{0pt}
	\setlength{\fboxrule}{0pt}
	\framebox[0.3\linewidth]{\footnotesize\Magg}
	\hspace{0.014\linewidth}
	\framebox[0.3\linewidth]{\footnotesize\Mdis}
    \hspace{0.014\linewidth}
	\framebox[0.3\linewidth]{\footnotesize\Mher}
	\\
	\vspace{0.015\linewidth}
	\setlength{\fboxrule}{0.1pt}
	\fbox{\includegraphics[trim={3.15cm 0 3.15cm 0},clip,width=0.3\linewidth]{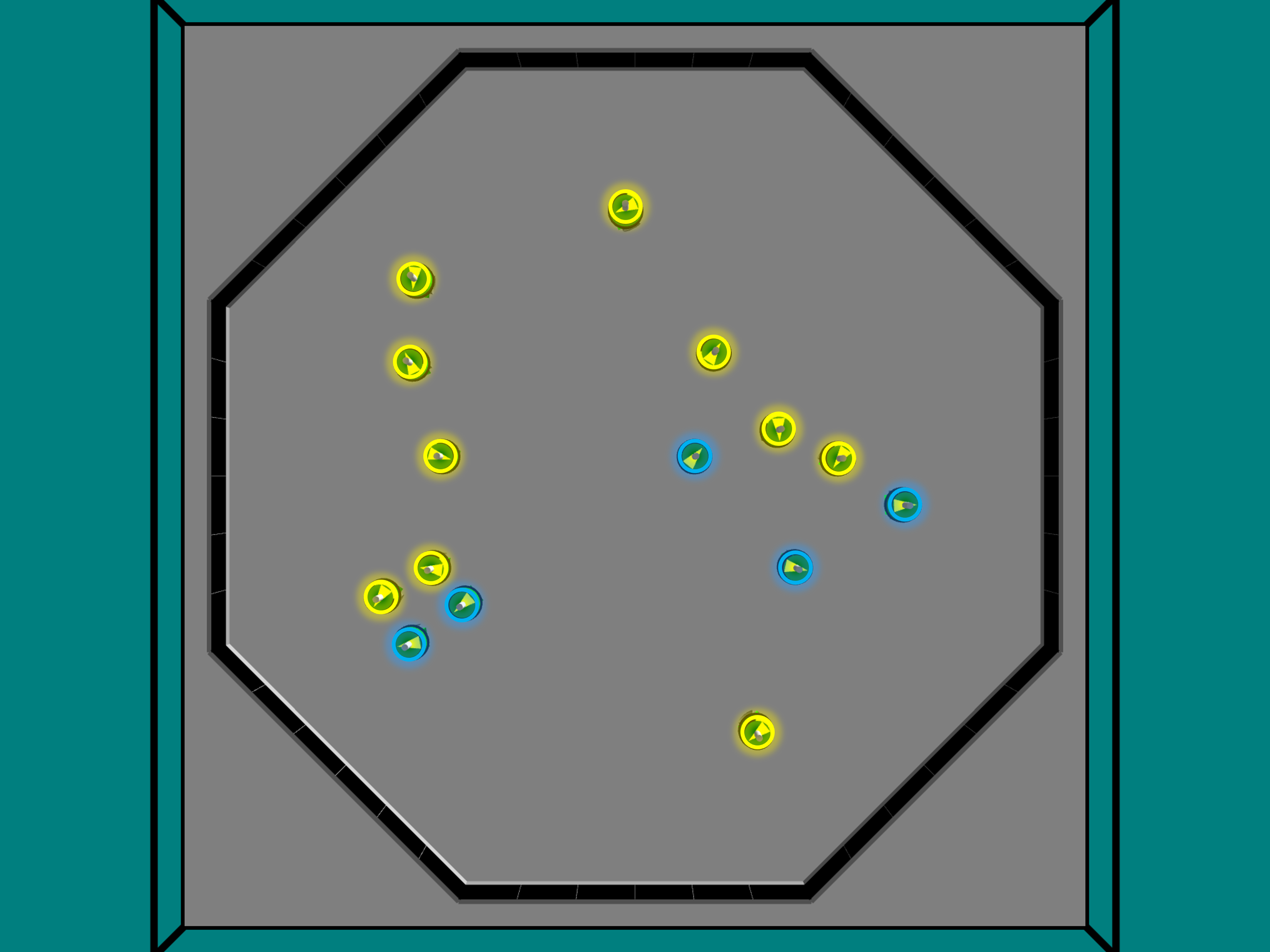}}
	\hspace{0.014\linewidth}
	\fbox{\includegraphics[trim={3.15cm 0 3.15cm 0},clip,width=0.3\linewidth]{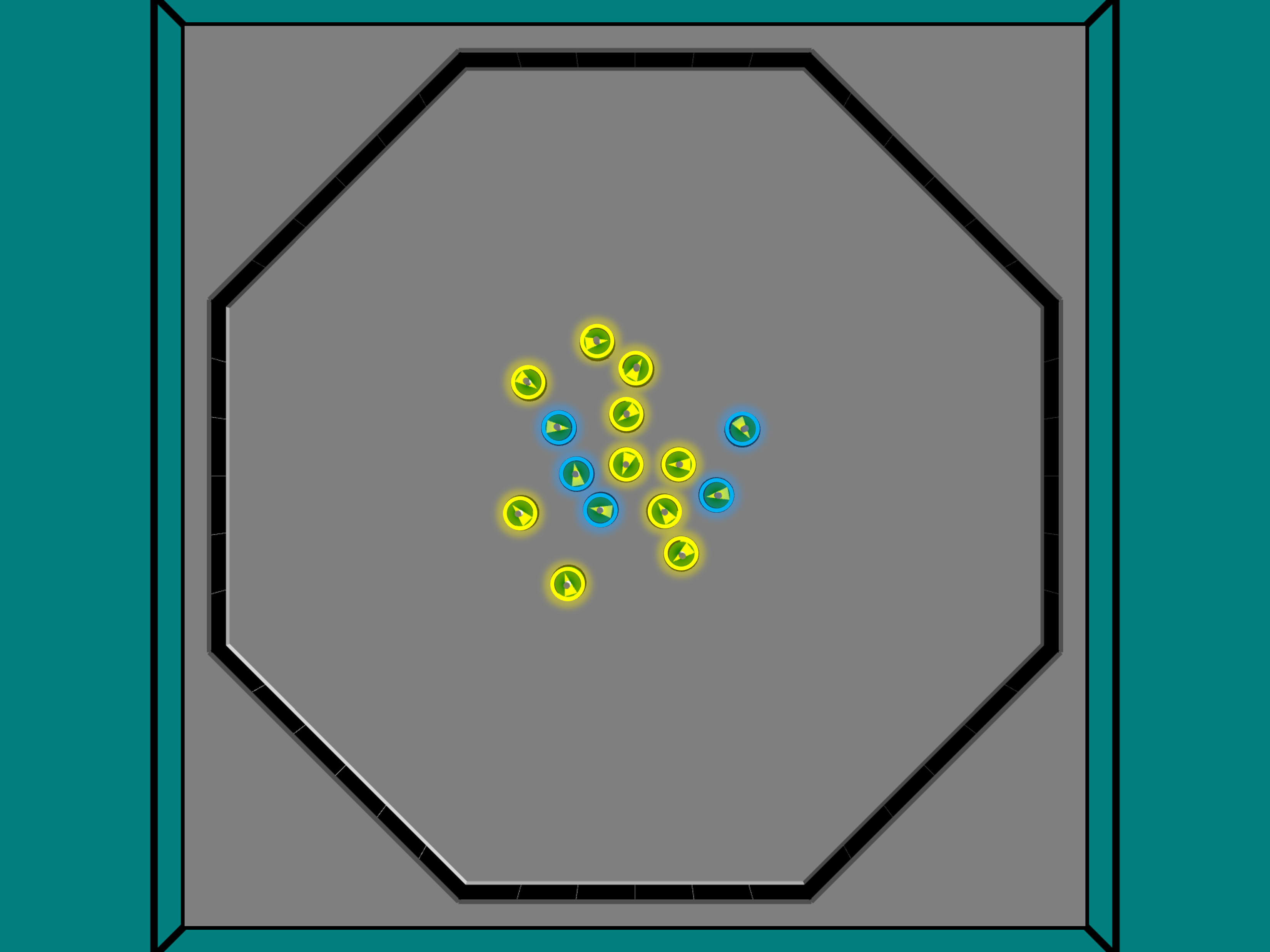}}
    \hspace{0.014\linewidth}
    \fbox{\includegraphics[trim={3.15cm 0 3.15cm 0},clip,width=0.3\linewidth]{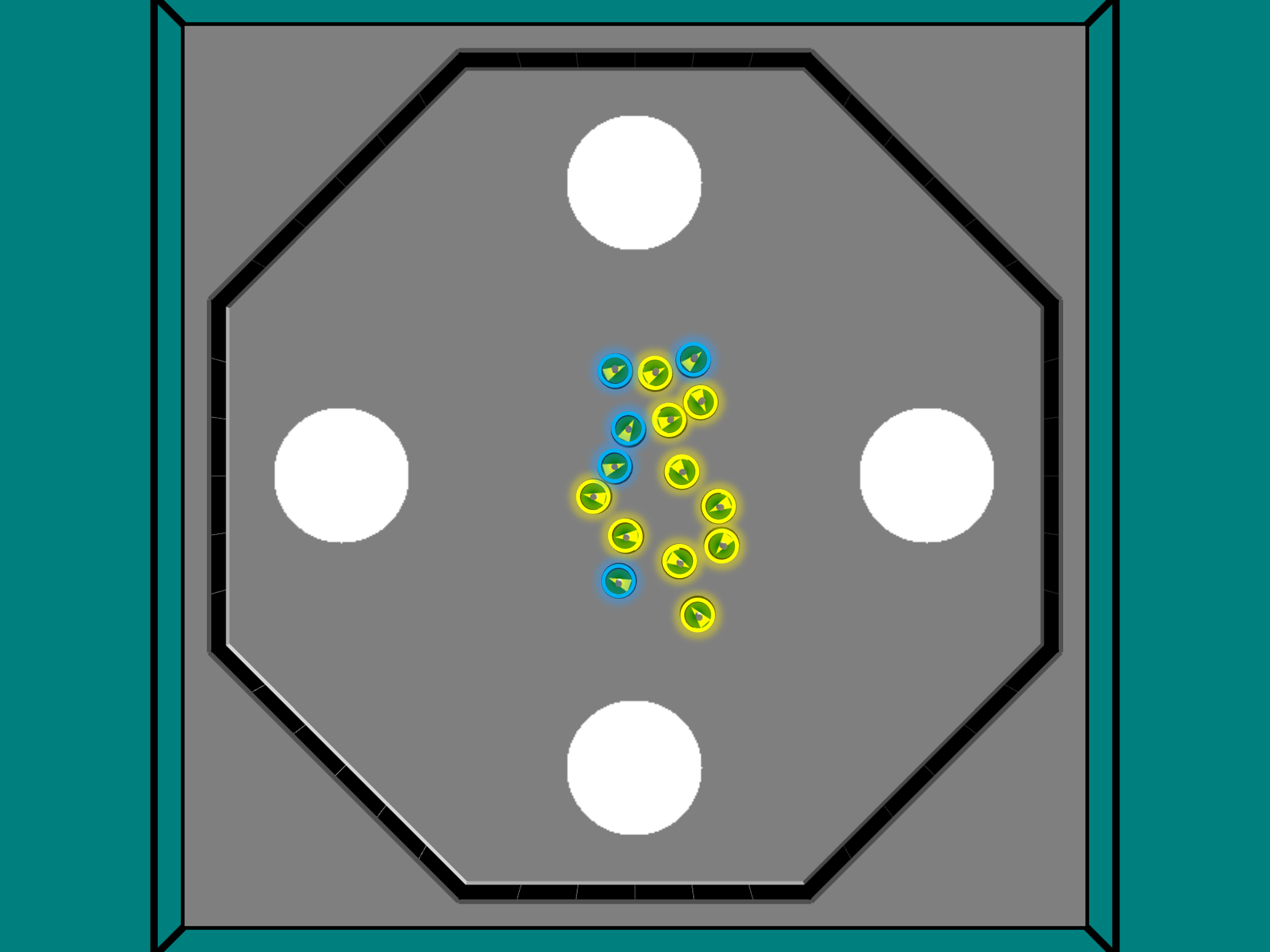}}
	
	\caption{
    \label{fig:missions} Experimental arenas for \Magg, \Mdis, and \Mher. The figure shows an example of the starting positions of five shepherds (cyan) and ten sheep (yellow) in each mission.}
\end{figure}

The robots operated in an octagonal arena of about $\SI{2.8}{\square\meter}$ and gray floor.
The arena had white floor regions on a per-mission basis---see Fig.~\ref{fig:missions}.
The robots had a time $T=\SI{120}{\second}$ to perform a mission.

\subsubsection{\Magg}
Shepherds and sheep are randomly distributed at the arena.
The shepherds must group the sheep.
The objective function $F_{1} = \frac{\sum_{i=1}^{10}\mathit{D}_{i}(T)}{10}$, which must be minimized, is the average distance from each sheep to the center of mass of all sheep at the end of the mission.
$\mathit{D}_{i}(T)$ is the distance from a sheep $i$ at position $(x_i,y_i)$ to the center of mass of all sheep $(x_c,y_c)$ at time~$T$.

\subsubsection{\Mdis}
Shepherds and sheep are randomly distributed at the center of the arena.
The shepherds must separate the sheep.
The objective function $F_{1}$, which in this case must be maximized, measures the average distance from each sheep to the center of mass of all sheep at the end of the mission.

\subsubsection{\Mher}
Shepherds and sheep are randomly distributed at the center of the arena.
The shepherds must drive the sheep to four indicated locations.
The indicated locations are white circular regions of about $\SI{0.3}{\square\meter}$.
The four locations are equivalent to each other.
The objective function $F_{2} = K_{\mathrm{n}}(T)$, which must be minimized, is the number of sheep that remain out of the four locations at the end of the mission, at time~$T$.

By pairing sheep control software and missions, we presented varied challenges to the design of the shepherding behaviors.
The sheep could be more or less cooperative with the shepherds for the mission at hand.
For example, we expected the shepherds to group the sheep more effectively when the sheep operated with \Batra, and less effectively when they operated with \Brepu.
Conversely, we expected the shepherds to separate the sheep more effectively when the sheep operated with \Brepu, and less effectively when they operated with \Batra.
\Batre gives more freedom to the automatic design process to select the best performing strategy.

\subsection{Baseline}
We also report results obtained with manual design and a random walk.
These results are a reference baseline to appraise the performance of the automatic methods.

\subsubsection{\chuma}
a manual design method introduced by \etal{Francesca}~\cite{FraBraBru-etal2014ants} to appraise the performance of automatic modular design methods.
In \chuma, a group of human designers fine-tune and combine parametric software modules to produce control software for the robots~\cite{FraBraBru-etal2014ants,FraBraBru-etal2015SI,KucVanBir2021evoapps}.
In our experiments, they produced control software for the shepherds.
We invited three designers with more than one year of experience in swarm robotics and some familiarity with \auto, the \epuck, and \argos.
The designers were provided with \pist's software modules and they were restricted to produce control software with its control architecture.
In this sense, the designers adopted the role of an optimization algorithm.
They searched the space of possible control software to generate the desired shepherding behaviors.

\subsubsection{\rwalk}
a trivial implementation of a random walk behavior in which the shepherds move with a ballistic motion~\cite{KegGarBir2019taros}.
\rwalk is not a design method, as no optimization process is conducted.
We include it as a lower bound to the performance, as in recent automatic design studies~\cite{HasLigRudBir2021NATUCOM,SalGarBir2024COMMENG}.

\subsection{Protocol}

We conducted experiments that pair the three missions and the three instances of sheep control software.
The pairing resulted in nine experimental scenarios.
Neither the automatic methods nor the human designers had direct access to information about the sheep control software.
The dynamics between shepherds and sheep had to be discovered during the design process via simulations.
The experimental protocol we followed is based on previous studies on the automatic design of robot swarms~\cite{FraBraBru-etal2014ants,FraBraBru-etal2015SI,KucVanBir2021evoapps}.
\pist and \evcmy were given a budget of $\num{100000}$ simulations to produce every instance of control software.
Producing control software with \pist and \evcmy costs less than producing it with \chuma.
Therefore, we used them to repeatedly produce more instances of control software than what \chuma could produce.
The number of evaluations was adjusted to obtain an equivalent number of observations for statistical analysis.
We produced 90 instances of control software with \pist and other 90 with \evcmy---10 per scenario.
Each of these instances was assessed once to obtain 90 observations per method.
\chuma produced 9 instances of control software, 1 per scenario.
We obtained the equivalent 90 observations by assessing each of these instances 10 times.
\rwalk was assessed 10 times in each scenario to obtain 90 observations.

We report numerical results with notched box-plots.
Additionally, we use a Friedman test to report aggregate results across the nine experimental scenarios.
Comparisons between methods are statistically supported by the computation of the $\SI{95}{\percent}$ confidence interval~\cite{Con1999book}.
The performance of two methods is considered statistically different if the confidence intervals of their median do not overlap.

\section{EXPERIMENTAL RESULTS} \label{sec:results}
\begin{figure}
    \vspace*{0.15cm}
	\centering
	\noindent
\resizebox{0.44\textwidth}{!}{
\begin{tikzpicture}[x=1pt,y=1pt]
\definecolor{fillColor}{RGB}{255,255,255}
\path[use as bounding box,fill=fillColor,fill opacity=0.00] (0,0) rectangle (399.13,119.13);
\begin{scope}[shift={(10.0,-10.0)}]
\begin{scope}
\path[clip] (  0.00,  0.00) rectangle (399.13,129.13);
\definecolor{drawColor}{RGB}{255,255,255}
\definecolor{fillColor}{RGB}{255,255,255}

\path[draw=drawColor,line width= 0.6pt,line join=round,line cap=round,fill=fillColor] (  0.00,  0.00) rectangle (399.13,129.13);
\end{scope}
\begin{scope}
\path[clip] ( 73.47, 42.30) rectangle (393.63,123.63);
\definecolor{fillColor}{RGB}{255,255,255}

\path[fill=fillColor] ( 73.47, 42.30) rectangle (393.63,123.63);
\definecolor{drawColor}{RGB}{211,211,211}

\path[draw=drawColor,line width= 0.2pt,line join=round] (291.61, 42.30) -- (291.61,123.63);

\path[draw=drawColor,line width= 0.2pt,line join=round] (111.03, 42.30) -- (111.03,123.63);

\path[draw=drawColor,line width= 0.2pt,line join=round] (379.07, 42.30) -- (379.07,123.63);

\path[draw=drawColor,line width= 0.2pt,line join=round] (114.98, 42.30) -- (114.98,123.63);

\path[draw=drawColor,line width= 0.2pt,line join=round] (268.60, 42.30) -- (268.60,123.63);

\path[draw=drawColor,line width= 0.2pt,line join=round] ( 88.02, 42.30) -- ( 88.02,123.63);

\path[draw=drawColor,line width= 0.2pt,line join=round] (356.06, 42.30) -- (356.06,123.63);

\path[draw=drawColor,line width= 0.2pt,line join=round] ( 91.97, 42.30) -- ( 91.97,123.63);
\definecolor{drawColor}{RGB}{0,0,0}

\path[draw=drawColor,line width= 0.6pt,line join=round] (291.61, 87.81) --
	(291.61, 97.49);

\path[draw=drawColor,line width= 0.6pt,line join=round] (291.61, 92.65) --
	(268.60, 92.65);

\path[draw=drawColor,line width= 0.6pt,line join=round] (268.60, 87.81) --
	(268.60, 97.49);

\path[draw=drawColor,line width= 0.6pt,line join=round] (111.03, 68.44) --
	(111.03, 78.12);

\path[draw=drawColor,line width= 0.6pt,line join=round] (111.03, 73.28) --
	( 88.02, 73.28);

\path[draw=drawColor,line width= 0.6pt,line join=round] ( 88.02, 68.44) --
	( 88.02, 78.12);

\path[draw=drawColor,line width= 0.6pt,line join=round] (379.07,107.17) --
	(379.07,116.85);

\path[draw=drawColor,line width= 0.6pt,line join=round] (379.07,112.01) --
	(356.06,112.01);

\path[draw=drawColor,line width= 0.6pt,line join=round] (356.06,107.17) --
	(356.06,116.85);

\path[draw=drawColor,line width= 0.6pt,line join=round] (114.98, 49.08) --
	(114.98, 58.76);

\path[draw=drawColor,line width= 0.6pt,line join=round] (114.98, 53.92) --
	( 91.97, 53.92);

\path[draw=drawColor,line width= 0.6pt,line join=round] ( 91.97, 49.08) --
	( 91.97, 58.76);
\definecolor{fillColor}{RGB}{0,0,0}

\path[draw=drawColor,line width= 0.8pt,line join=round,line cap=round,fill=fillColor] (280.11, 92.65) circle (  2.85);

\path[draw=drawColor,line width= 0.8pt,line join=round,line cap=round,fill=fillColor] ( 99.52, 73.28) circle (  2.85);

\path[draw=drawColor,line width= 0.8pt,line join=round,line cap=round,fill=fillColor] (367.57,112.01) circle (  2.85);

\path[draw=drawColor,line width= 0.8pt,line join=round,line cap=round,fill=fillColor] (103.48, 53.92) circle (  2.85);

\path[draw=drawColor,line width= 1.1pt,line join=round,line cap=round] ( 73.47, 42.30) rectangle (393.63,123.63);
\end{scope}
\begin{scope}
\definecolor{drawColor}{RGB}{0,0,0}

\node[text=drawColor,rotate= 0.00,anchor=base ,inner sep=0pt, outer sep=2pt, scale=  1.2] at ( 365.90, 58.60) {The lower,};

\node[text=drawColor,rotate= 0.00,anchor=base ,inner sep=0pt, outer sep=2pt, scale=  1.2] at ( 365.90, 48.60) {the better.};

\node[text=drawColor,anchor=base east,inner sep=0pt, outer sep=0pt, scale=  1.4] at ( 64.15, 50.89) {\texttt{Pistacchio}};

\node[text=drawColor,anchor=base east,inner sep=0pt, outer sep=0pt, scale=  1.4] at ( 64.15, 70.25) {\texttt{EvoCMY}};

\node[text=drawColor,anchor=base east,inner sep=0pt, outer sep=0pt, scale=  1.4] at ( 64.15, 89.62) {\texttt{C-Human}};

\node[text=drawColor,anchor=base east,inner sep=0pt, outer sep=0pt, scale=  1.4] at ( 64.15,108.98) {\texttt{R-Walk}};
\end{scope}
\begin{scope}
\path[clip] (  0.00,  0.00) rectangle (399.13,129.13);
\definecolor{drawColor}{gray}{0.20}

\path[draw=drawColor,line width= 0.6pt,line join=round] ( 66.35, 53.92) --
	( 73.47, 53.92);

\path[draw=drawColor,line width= 0.6pt,line join=round] ( 66.35, 73.28) --
	( 73.47, 73.28);

\path[draw=drawColor,line width= 0.6pt,line join=round] ( 66.35, 92.65) --
	( 73.47, 92.65);

\path[draw=drawColor,line width= 0.6pt,line join=round] ( 66.35,112.01) --
	( 73.47,112.01);
\end{scope}
\begin{scope}
\path[clip] (  0.00,  0.00) rectangle (399.13,129.13);
\definecolor{drawColor}{gray}{0.20}

\path[draw=drawColor,line width= 0.6pt,line join=round] ( 78.11, 35.19) --
	( 78.11, 42.30);

\path[draw=drawColor,line width= 0.6pt,line join=round] (142.19, 35.19) --
	(142.19, 42.30);

\path[draw=drawColor,line width= 0.6pt,line join=round] (206.26, 35.19) --
	(206.26, 42.30);

\path[draw=drawColor,line width= 0.6pt,line join=round] (270.34, 35.19) --
	(270.34, 42.30);

\path[draw=drawColor,line width= 0.6pt,line join=round] (334.41, 35.19) --
	(334.41, 42.30);
\end{scope}
\begin{scope}
\path[clip] (  0.00,  0.00) rectangle (399.13,129.13);
\definecolor{drawColor}{RGB}{0,0,0}

\node[text=drawColor,anchor=base,inner sep=0pt, outer sep=0pt, scale=  1.4] at ( 78.11, 23.93) {10};

\node[text=drawColor,anchor=base,inner sep=0pt, outer sep=0pt, scale=  1.4] at (142.19, 23.93) {15};

\node[text=drawColor,anchor=base,inner sep=0pt, outer sep=0pt, scale=  1.4] at (206.26, 23.93) {20};

\node[text=drawColor,anchor=base,inner sep=0pt, outer sep=0pt, scale=  1.4] at (270.34, 23.93) {25};

\node[text=drawColor,anchor=base,inner sep=0pt, outer sep=0pt, scale=  1.4] at (334.41, 23.93) {30};
\end{scope}
\begin{scope}
\path[clip] (  0.00,  0.00) rectangle (399.13,129.13);
\definecolor{drawColor}{RGB}{0,0,0}

\node[text=drawColor,anchor=base,inner sep=0pt, outer sep=0pt, scale=  1.6] at (233.55,  4.64) {Average rank};
\end{scope}
\end{scope}
\end{tikzpicture}
}
	\caption{
    \label{fig:friedman} Friedman test that aggregates the results obtained in the nine experimental scenarios. The plot shows the average rank of each method and its confidence interval.}
\end{figure}
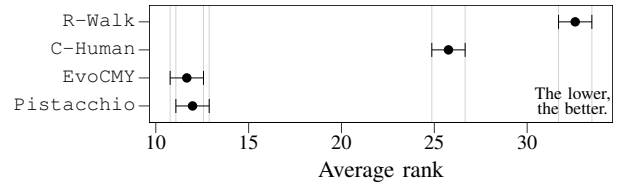

The control software, collected data, and demonstrative videos are available as supplementary material~\cite{GarBir2024icra-supp}.
\begin{figure}
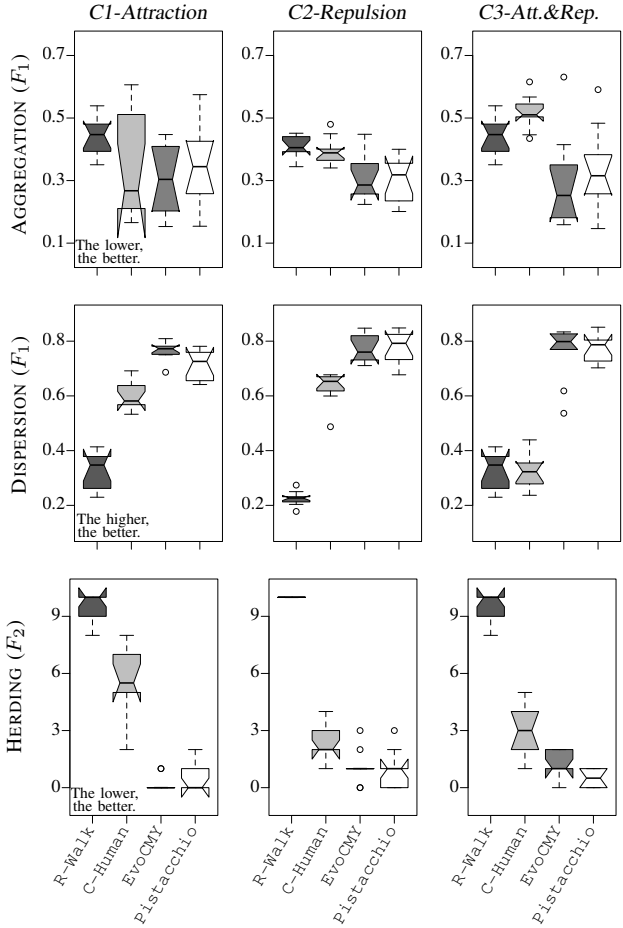

	\centering
    \rotatebox[origin=l]{90}{\hspace{0.60cm}{\footnotesize\Magg~($F_1$)}}
	\noindent\input{img/pst_ct.tex}
    \rotatebox[origin=l]{90}{\hspace{0.8cm}{\footnotesize\Mdis~($F_1$)}}
    \noindent\input{img/pst_ds.tex}
    \rotatebox[origin=l]{90}{\hspace{2.2cm}{\footnotesize\Mher~($F_2$)}}
    \noindent\input{img/pst_sc.tex}
	\caption{
    \label{fig:plots} Results per mission and sheep control software. The plots show the score obtained in the nine experimental scenarios, ten observations per method and scenario. Results per mission are organized in rows. Results per sheep control software are organized in columns. Results per design method are presented with grayscale box-plots, \rwalk~(~\raisebox{2pt}{\fcolorbox{black}{black!75!white}{\rule{0pt}{1pt}\rule{1pt}{0pt}}}~), \chuma~(~\raisebox{2pt}{\fcolorbox{black}{black!35!white}{\rule{0pt}{1pt}\rule{1pt}{0pt}}}~), \evcmy~(~\raisebox{2pt}{\fcolorbox{black}{black!50!white}{\rule{0pt}{1pt}\rule{1pt}{0pt}}}~), \pist~(~\raisebox{2pt}{\fcolorbox{black}{white}{\rule{0pt}{1pt}\rule{1pt}{0pt}}}~).}
\end{figure}
Fig.~\ref{fig:plots} shows box-plots of the score obtained in each mission.
Fig.~\ref{fig:friedman} shows the average rank of the design methods over the nine experimental scenarios.
The per-mission results and the Friedman test did not detect any significant difference between the performance of \pist and \evcmy, but the two are significantly better than \chuma and \rwalk.
Moreover, \chuma was significantly better than \rwalk.
The results show that automatic design was more effective than manual design in addressing the shepherding problems we considered.
Also, all design methods generated collective behaviors that are more effective than the simple random walk---the lower bound.
Our simulation-only comparison between \pist and \evcmy was not sufficient to identify possible performance differences between the modular and the neuroevolutionary approach---which have been previously reported in similar studies~\cite{FraBraBru-etal2015SI,LigBir2020SI}.

\subsection{Shepherding strategies}

\pist and \evcmy leveraged coordination and localization via color signals to create effective interactions between shepherds and sheep.
In most cases, the two methods designed behaviors in which the shepherds stimulated the sheep in similar ways.
We describe below some shepherding strategies that were generated.
The following are general observations we made over all instances of control software.

\subsubsection{Grouping the sheep in \Magg}
When the sheep operated with \Batra, the shepherds displayed magenta to attract them and they were themselves also attracted to other shepherds that displayed magenta.
In this way, shepherds and sheep remained close to each other---keeping the sheep close to their center of mass.
The automatic methods designed a coordinated cooperative behavior between shepherds.
When the sheep operated with a \Brepu, the shepherds displayed cyan and moved on a circular trajectory close to the walls of the arena.
The sheep were therefore steadily repelled towards the center of the arena and formed a single group.
When the sheep operated with \Batre, the shepherds used a behavior similar to that observed in \Batra or \Brepu---no noticeable preference was observed.

\subsubsection{Separating the sheep in \Mdis}
When the sheep operated with \Batra, the shepherds remained close to the walls and displayed magenta to attract the sheep.
This behavior dispersed the sheep along the edges of the arena---keeping them far from their center of mass.
When the sheep operated with \Brepu, the shepherds moved in circles in the center of the arena while displaying cyan.
In this way, the shepherds separated the sheep by steadily pushing them towards the walls.
When the sheep operated with \Batre, the shepherds used a behavior similar to that observed in \Batra or \Brepu---also no noticeable preference was observed.

\subsubsection{Driving the sheep in \Mher}
The shepherds reacted to the color yellow that the sheep displayed.
When the sheep operated with \Batra, the shepherds displayed magenta to attract the sheep.
Simultaneously, they were also repelled from the color yellow the sheep displayed.
In this way, the shepherds guided the sheep from the front.  
The two navigated the arena together until the sheep stepped into a white region and turned off their LEDs.
When the sheep operated with \Brepu, the shepherds displayed cyan to repel the sheep.
Unlike the behavior observed in \Batra, in this case, the shepherds were attracted to the color yellow that the sheep displayed.
The simultaneous execution of these two behaviors resulted in shepherds that chased the sheep until the latter stepped into a white region and turned off their LEDs.
The shepherds used a similar behavior to that observed in \Batra or \Brepu when the sheep operated with \Batre.
Also in this case, no noticeable preference was observed.

Color signaling and physical proximity were two alternative ways for the shepherds to interact with the sheep.
Unlike \pist and \evcmy, \chuma produced control software that leveraged the color displayed by the shepherds only in a few cases.
\chuma and \pist operate on the same set of modules and software architecture, and can potentially produce control software that performs similarly.
However, the human designers mainly used sub-optimal strategies (based on physical proximity) to enable the interaction between shepherds and sheep.
In a way, they failed to discover and use the most effective strategy: the interaction via colors.

\section{DISCUSSION}

The automatic methods effectively searched the design space and exploited the dynamics between shepherd and sheep.
The shepherds used color stimuli to interact with the sheep in a meaningful and good-performing way.
In some cases, the shepherds coordinated with each other using the same stimuli.
Past research has shown that automatic design and simple color signaling are sufficient to generate mission-specific coordination and spatial-organization strategies~\cite{GarBir2020AS,SalGarBir2024COMMENG}.
These results were obtained with a homogeneous robot swarm.
In our experiments, \pist and \evcmy generated similar mission-specific coordination and spatial organization between shepherds and sheep.
We show therefore that this ability of automatic methods extends to heterogeneous setups.

So far, automatic design research has focused mainly on missions performed by a single robot swarm that operates alone.
With our experiments, we showed that existing approaches are also well suited to address more complex heterogeneous scenarios.
We consider heterogeneity in both the number of swarms that populate an environment and the type of control software they use.
In our experiments, the sheep operated with finite-state machines.
\pist and \evcmy generated effective shepherding strategies despite that their shepherds operated with different kinds of control software---finite-state machines and neural networks, respectively.
A key element to enable a setup with heterogeneous control software was to formally define the shepherd and sheep robot capabilities within a single reference model, \RMD{3}.

\emph{A priori}, we expected to observe a significant performance difference in \Magg and \Mdis when the sheep operated with diametrically different behaviors such as \Batra and \Brepu.
However, no such performance difference was observed.
This indicates that the automatic methods simultaneously tailored the control software of the shepherds to the one of the sheep, and to the goal of the mission---regardless of the combination.
On the other hand, \chuma was less effective.
Human designers had difficulties on exploring the design space and finding good-performing shepherding strategies.
We conceived the experiments in a way that the dynamics between shepherd and sheep had to be discovered during the design process.
Automatic methods have a notable advantage in this task.
The optimization process is more effective than a human designer in exploring the large and complex design space.

The shepherding problem was an appropriate framework to study the design of robot swarms that must interact with other robots.
Our current experimental setup can be directly extended to missions that involve other types of interactions.
The sheep we considered are rather individualistic:
they react individually to the stimuli of the shepherds without considering the behavior of other sheep (beyond the physical proximity).
Moreover, their naturally-static behavior made them easy to handle for shepherds.
One could possibly create more complex missions with sheep that continuously move and operate with a more coordinated collective behavior---both cooperative or adversarial.
In this sense, our research could bootstrap recent studies on the design of robot swarms that are robust to attacks from adversarial robots~\cite{StrCasDor2020FRAI,CasHarPenDor2021SCIROB,StrPacDor2023SCIROB,ReiZakDemFer2023COMPHY}.

\section{CONCLUSION} \label{sec:conclusions}

Automatic design is a viable approach to producing swarms that operate in environments populated by other robots. 
We showed this with experiments conducted in a robot shepherding problem.
The automatic methods designed control software for the shepherds, which effectively coordinated groups of pre-programmed sheep.
The automatic methods leveraged mission-specific coordination and spatial organization to generate good-performing control software.
These capabilities were shown in the past in the automatic design of homogeneous robot swarms.
We showed they also hold for an heterogeneous system that must operate collectively---the shepherds and sheep.

In our experiments, the sheep operated with finite-state machines and the shepherds operated either with finite-state machines or artificial neural networks.
In this sense, we demonstrated the automatic design of collective behaviors between swarms that operate with different control architecture.
We did so by experimenting with the modular and neuroevolutionary approaches.
We expect these experiments can motivate new research on the automatic design of heterogeneous robot swarms---with multiple different robot swarms designed at once.
We will further investigate this idea with the simultaneous design of shepherds and sheep.
%








\IEEEtriggeratref{49}
\bibliographystyle{IEEEtran}
\bibliography{demiurge-bib/definitions.bib,demiurge-bib/author.bib,demiurge-bib/address.bib,demiurge-bib/proceedings.bib,demiurge-bib/journal.bib,demiurge-bib/publisher.bib,demiurge-bib/series.bib,demiurge-bib/institution.bib,demiurge-bib/bibliography.bib,demiurge-bib/newbibliography.bib}

\begin{thebibliography}{10}
\providecommand{\url}[1]{#1}
\csname url@rmstyle\endcsname
\providecommand{\newblock}{\relax}
\providecommand{\bibinfo}[2]{#2}
\providecommand\BIBentrySTDinterwordspacing{\spaceskip=0pt\relax}
\providecommand\BIBentryALTinterwordstretchfactor{4}
\providecommand\BIBentryALTinterwordspacing{\spaceskip=\fontdimen2\font plus
\BIBentryALTinterwordstretchfactor\fontdimen3\font minus
  \fontdimen4\font\relax}
\providecommand\BIBforeignlanguage[2]{{%
\expandafter\ifx\csname l@#1\endcsname\relax
\typeout{** WARNING: IEEEtran.bst: No hyphenation pattern has been}%
\typeout{** loaded for the language `#1'. Using the pattern for}%
\typeout{** the default language instead.}%
\else
\language=\csname l@#1\endcsname
\fi
#2}}

\bibitem{Ham2018book}
H.~Hamann, \emph{Swarm robotics: a formal approach}.\hskip 1em plus 0.5em minus
  0.4em\relax Cham, Switzerland: Springer, 2018.

\bibitem{Sah2005sab}
E.~{\c S}ahin, ``Swarm robotics: from sources of inspiration to domains of
  application,'' in \emph{Swarm Robotics: SAB 2004 International Workshop},
  ser. Lecture Notes in Computer Science, E.~{\c S}ahin and W.~M. Spears, Eds.,
  vol. 3342.\hskip 1em plus 0.5em minus 0.4em\relax Berlin, Germany: Springer,
  2005, pp. 10--20.

\bibitem{Ben2005sab}
G.~Beni, ``From swarm intelligence to swarm robotics,'' in \emph{Swarm
  Robotics: SAB 2004 International Workshop}, ser. Lecture Notes in Computer
  Science, E.~{\c S}ahin and W.~M. Spears, Eds., vol. 3342.\hskip 1em plus
  0.5em minus 0.4em\relax Berlin, Germany: Springer, 2005, pp. 1--9.

\bibitem{BraFerBirDor2013SI}
M.~Brambilla, E.~Ferrante, M.~Birattari, and M.~Dorigo, ``Swarm robotics: a
  review from the swarm engineering perspective,'' \emph{Swarm Intelligence},
  vol.~7, no.~1, pp. 1--41, 2013.

\bibitem{StoVarSvoBel2020FRAI}
D.~St-Onge, V.~S. Varadharajan, {\v S}.~Ivan, and G.~Beltrame, ``From design to
  deployment: decentralized coordination of heterogeneous robotic teams,''
  \emph{Frontiers in Robotics and AI}, vol.~7, p.~51, 2020.

\bibitem{BirLigBoz-etal2019FRAI}
M.~Birattari, A.~Ligot, D.~Bozhinoski, M.~Brambilla, G.~Francesca,
  L.~Garattoni, D.~Garzón~Ramos, K.~Hasselmann, M.~Kegeleirs, J.~Kuckling,
  F.~Pagnozzi, A.~Roli, M.~Salman, and T.~Stützle, ``Automatic off-line design
  of robot swarms: a manifesto,'' \emph{Frontiers in Robotics and AI}, vol.~6,
  p.~59, 2019.

\bibitem{BirLigHas2020NATUMINT}
M.~Birattari, A.~Ligot, and K.~Hasselmann, ``Disentangling automatic and
  semi-automatic approaches to the optimization-based design of control
  software for robot swarms,'' \emph{Nature Machine Intelligence}, vol.~2,
  no.~9, pp. 494--499, 2020.

\bibitem{FraBir2016FRAI}
G.~Francesca and M.~Birattari, ``Automatic design of robot swarms: achievements
  and challenges,'' \emph{Frontiers in Robotics and AI}, vol.~3, no.~29, pp.
  1--9, 2016.

\bibitem{LieBaySow-etal2004icra}
J.-M. Lien, O.~B. Bayazit, R.~T. Sowell, S.~Rodríguez, and N.~M. Amato,
  ``Shepherding behaviors,'' in \emph{2004 IEEE International Conference on
  Robotics and Automation (ICRA)}, vol.~4.\hskip 1em plus 0.5em minus
  0.4em\relax Piscataway, NJ, USA: IEEE, 2004, pp. 4159--4164.

\bibitem{SchUmlSenElm2020FRAI}
M.~Schranz, M.~Umlauft, M.~Sende, and W.~Elmenreich, ``Swarm robotic behaviors
  and current applications,'' \emph{Frontiers in Robotics and AI}, vol.~7,
  p.~36, 2020.

\bibitem{NolFlo2000book}
S.~Nolfi and D.~Floreano, \emph{Evolutionary Robotics: The Biology,
  Intelligence, and Technology of Self-Organizing Machines}, 1st~ed.\hskip 1em
  plus 0.5em minus 0.4em\relax Cambridge, MA, USA: MIT Press, 2000, a Bradford
  Book.

\bibitem{Tri2008book}
V.~Trianni, \emph{Evolutionary Swarm Robotics}.\hskip 1em plus 0.5em minus
  0.4em\relax Berlin, Germany: Springer, 2008.

\bibitem{Nol2021book}
S.~Nolfi, \emph{Behavioral and Cognitive Robotics: An Adaptive
  Perspective}.\hskip 1em plus 0.5em minus 0.4em\relax Rome, Italy: Institute
  of Cognitive Sciences and Technologies, National Research Council, 2021.

\bibitem{FraBraBru-etal2014SI}
G.~Francesca, M.~Brambilla, A.~Brutschy, V.~Trianni, and M.~Birattari,
  ``{AutoMoDe}: a novel approach to the automatic design of control software
  for robot swarms,'' \emph{Swarm Intelligence}, vol.~8, no.~2, pp. 89--112,
  2014.

\bibitem{BirLigFra2021admlsa}
M.~Birattari, A.~Ligot, and G.~Francesca, ``{AutoMoDe}: a modular approach to
  the automatic off-line design and fine-tuning of control software for robot
  swarms,'' in \emph{Automated Design of Machine Learning and Search
  Algorithms}, ser. Natural Computing Series, N.~Pillay and R.~Qu, Eds.\hskip
  1em plus 0.5em minus 0.4em\relax Cham, Switzerland: Springer, 2021, pp.
  73--90.

\bibitem{ChrDor2006alife}
A.~L. Christensen and M.~Dorigo, ``Evolving an integrated phototaxis and
  hole-avoidance behavior for a swarm-bot,'' in \emph{Artificial Life X:
  Proceedings of the Tenth International Conference on the Simulation and
  Synthesis of Living Systems}, L.~M. Rocha, L.~S. Yaeger, M.~A. Bedau,
  D.~Floreano, R.~L. Goldstone, and A.~Vespignani, Eds.\hskip 1em plus 0.5em
  minus 0.4em\relax Cambridge, MA, USA: MIT Press, 2006, pp. 248--254, a
  Bradford Book.

\bibitem{JonWinHauStu2019AIS}
S.~Jones, A.~Winfield, S.~Hauert, and M.~Studley, ``Onboard evolution of
  understandable swarm behaviors,'' \emph{Advanced Intelligent Systems},
  vol.~1, no.~6, p. 1900031, 2019.

\bibitem{GhaKucGarBir2023icra-up}
I.~Gharbi, J.~Kuckling, D.~Garzón~Ramos, and M.~Birattari, ``Show me what you
  want: inverse reinforcement learning to automatically design robot swarms by
  demonstration,'' in \emph{2023 IEEE International Conference on Robotics and
  Automation (ICRA)}.\hskip 1em plus 0.5em minus 0.4em\relax Piscataway, NJ,
  USA: IEEE, 2023, pp. 5063--5070.

\bibitem{FraBraBru-etal2015SI}
G.~Francesca, M.~Brambilla, A.~Brutschy, L.~Garattoni, R.~Miletitch,
  G.~Podevijn, A.~Reina, T.~Soleymani, M.~Salvaro, C.~Pinciroli, F.~Mascia,
  V.~Trianni, and M.~Birattari, ``{AutoMoDe-Chocolate}: automatic design of
  control software for robot swarms,'' \emph{Swarm Intelligence}, vol.~9, no.
  2--3, pp. 125--152, 2015.

\bibitem{CamFer2022gecco}
N.~Cambier and E.~Ferrante, ``{AutoMoDe}-{Pomodoro}: an evolutionary class of
  modular designs,'' in \emph{GECCO'22: Proceedings of the Genetic and
  Evolutionary Computation Conference}, J.~E. Fieldsend, Ed.\hskip 1em plus
  0.5em minus 0.4em\relax New York, NY, USA: ACM, 2022, pp. 100--103.

\bibitem{KucVanBir2021evoapps}
J.~Kuckling, V.~van Pelt, and M.~Birattari, ``Automatic modular design of
  behavior trees for robot swarms with communication capabilities,'' in
  \emph{Applications of Evolutionary Computation: 24th International
  Conference, EvoApplications 2021}, ser. Lecture Notes in Computer Science,
  P.~A. Castillo and J.~L. Jiménez~Laredo, Eds., vol. 12694.\hskip 1em plus
  0.5em minus 0.4em\relax Cham, Switzerland: Springer, 2021, pp. 130--145.

\bibitem{DuaCosGom-etal2016PLOSONE}
M.~Duarte, V.~Costa, J.~Gomes, T.~Rodrigues, F.~Silva, S.~M. Oliveira, and
  A.~L. Christensen, ``Evolution of collective behaviors for a real swarm of
  aquatic surface robots,'' \emph{PLOS ONE}, vol.~11, no.~3, p. e0151834, 2016.

\bibitem{GarBir2020AS}
D.~Garzón~Ramos and M.~Birattari, ``Automatic design of collective behaviors
  for robots that can display and perceive colors,'' \emph{Applied Sciences},
  vol.~10, no.~13, p. 4654, 2020.

\bibitem{SalGarBir2024COMMENG}
M.~Salman, D.~Garzón~Ramos, and M.~Birattari, ``Automatic design of
  stigmergy-based behaviours for robot swarms,'' \emph{Communications
  Engineering}, vol.~3, p.~30, 2024.

\bibitem{KegGarHas-etal2024RAL}
M.~Kegeleirs, D.~Garzón~Ramos, K.~Hasselmann, L.~Garattoni, G.~Francesca, and
  M.~Birattari, ``Transferability in the automatic off-line design of robot
  swarms: from sim-to-real to embodiment and design-method transfer across
  different platforms,'' \emph{IEEE Robotics and Automation Letters}, vol.~9,
  no.~3, pp. 2758--2765, 2024.

\bibitem{TriGroLab-etal2003ecal}
V.~Trianni, R.~Groß, H.~Labella, Thomas, E.~{\c S}ahin, and M.~Dorigo,
  ``Evolving aggregation behaviors in a swarm of robots,'' in \emph{Advances in
  Artificial Life: 7th European Conference, ECAL 2003}, ser. Lecture Notes in
  Computer Science, W.~Banzhaf, J.~Ziegler, T.~Christaller, P.~Dittrich, and
  J.~T. Kim, Eds., vol. 2801.\hskip 1em plus 0.5em minus 0.4em\relax Berlin,
  Germany: Springer, 2003, pp. 865--874.

\bibitem{FerTurDue-etal2015PLOSCB}
E.~Ferrante, A.~E. Turgut, E.~A. Duéñez-Guzmán, M.~Dorigo, and
  T.~Wenseleers, ``Evolution of self-organized task specialization in robot
  swarms,'' \emph{PLOS Computational Biology}, vol.~11, no.~8, p. e1004273,
  2015.

\bibitem{MenGarMor-etal2022SWEVO}
F.~J. Mendiburu, D.~Garzón~Ramos, M.~R.~A. Morais, A.~M.~N. Lima, and
  M.~Birattari, ``{AutoMoDe}-{Mate}: automatic off-line design of
  spatially-organizing behaviors for robot swarms,'' \emph{Swarm and
  Evolutionary Computation}, vol.~74, p. 101118, 2022.

\bibitem{HamSchElm-etal2020FRAI}
H.~Hamann, M.~Schranz, W.~Elmenreich, V.~Trianni, C.~Pinciroli, N.~Bredeche,
  and E.~Ferrante, ``Editorial: designing self-organization in the physical
  realm,'' \emph{Frontiers in Robotics and AI}, vol.~7, p. 164, 2020.

\bibitem{DorTheTri2020SCIROB}
M.~Dorigo, G.~Theraulaz, and V.~Trianni, ``Reflections on the future of swarm
  robotics,'' \emph{Science Robotics}, vol.~5, p. eabe4385, 2020.

\bibitem{DorTheTri2021PIEEE}
------, ``Swarm robotics: past, present, and future [point of view],''
  \emph{Proceedings of the IEEE}, vol. 109, no.~7, pp. 1152--1165, 2021.

\bibitem{KinPorStr-etal2023}
A.~J. King, S.~J. Portugal, D.~Strömbom, R.~P. Mann, J.~A. Carrillo,
  D.~Kalise, G.~de~Croon, H.~Barnett, P.~Scerri, R.~Groß, D.~R. Chadwick, and
  M.~Papadopoulou, ``Biologically inspired herding of animal groups by
  robots,'' \emph{Methods in Ecology and Evolution}, vol.~14, no.~2, pp.
  478--486, 2023.

\bibitem{GenSto2014ants}
K.~Genter and P.~Stone, ``Influencing a flock via ad hoc teamwork,'' in
  \emph{Swarm Intelligence: 9th International Conference, ANTS 2014}, ser.
  Lecture Notes in Computer Science, M.~Dorigo, M.~Birattari, S.~Garnier,
  H.~Hamann, M.~Montes~de Oca, C.~Solnon, and T.~Stützle, Eds., vol.
  8667.\hskip 1em plus 0.5em minus 0.4em\relax Cham, Switzerland: Springer
  International Publishing, 2014, pp. 110--121.

\bibitem{GenSto2016aamas}
------, ``Adding influencing agents to a flock,'' in \emph{AAMAS '16:
  Proceedings of the 2016 International Conference on Autonomous Agents and
  Multiagent Systems}.\hskip 1em plus 0.5em minus 0.4em\relax Richland, SC,
  USA: International Foundation for Autonomous Agents and Multiagent Systems
  (IFAAMAS), 2016, pp. 615--623.

\bibitem{OzdGauGro2017ecal}
A.~Özdemir, M.~Gauci, and R.~Groß, ``Shepherding with robots that do not
  compute,'' in \emph{ECAL 2017, the Fourteenth European Conference on
  Artificial Life}.\hskip 1em plus 0.5em minus 0.4em\relax Cambridge, MA, USA:
  MIT Press, 2017, pp. 332--339.

\bibitem{LicBelDix2019IEEETR}
R.~A. Licitra, Z.~I. Bell, and W.~E. Dixon, ``Single-agent indirect herding of
  multiple targets with uncertain dynamics,'' \emph{IEEE Transactions on
  Robotics}, vol.~35, no.~4, pp. 847--860, 2019.

\bibitem{PieSch2018IEEETR}
A.~Pierson and M.~Schwager, ``Controlling noncooperative herds with robotic
  herders,'' \emph{IEEE Transactions on Robotics}, vol.~34, no.~2, pp.
  517--525, 2018.

\bibitem{HuTurKra-etal2020IEEETCDS}
J.~Hu, A.~E. Turgut, T.~Krajník, B.~Lennox, and F.~Arvin, ``Occlusion-based
  coordination protocol design for autonomous robotic shepherding tasks,''
  \emph{IEEE Transactions on Cognitive and Developmental Systems}, p.~1, 2020.

\bibitem{SebMonSag2022IEEETR}
E.~Sebastián, E.~Montijano, and C.~Sagüés, ``Adaptive multirobot implicit
  control of heterogeneous herds,'' \emph{IEEE Transactions on Robotics},
  vol.~38, no.~6, pp. 3622--3635, 2022.

\bibitem{NedSil2019SWEVO}
N.~Nedjah and L.~Silva~Junior, ``Review of methodologies and tasks in swarm
  robotics towards standardization,'' \emph{Swarm and Evolutionary
  Computation}, vol.~50, p. 100565, 2019.

\bibitem{DosOzdGauGro2022ants}
G.~Y. Dosieah, A.~Özdemir, M.~Gauci, and R.~Groß, ``Moving mixtures
  of active and passive elements with robots that do not compute,'' in
  \emph{Swarm Intelligence: 13th International Conference, ANTS 2022}, ser.
  Lecture Notes in Computer Science, vol. 13491.\hskip 1em plus 0.5em minus
  0.4em\relax Cham, Switzerland: Springer, 2022, pp. 183--195.

\bibitem{GarBir2024icra-supp}
D.~Garzón~Ramos and M.~Birattari, ``Automatically designing robot swarms in
  environments populated by other robots, an experiment in robot shepherding:
  supplementary material,''
  \url{https://iridia.ulb.ac.be/supp/IridiaSupp2023-002}, 2024.

\bibitem{MonBonRae-etal2009arsc}
F.~Mondada, M.~Bonani, X.~Raemy, J.~Pugh, C.~Cianci, A.~Klaptocz, S.~Magnenat,
  J.-C. Zufferey, D.~Floreano, and A.~Martinoli, ``The e-puck, a robot designed
  for education in engineering,'' in \emph{ROBOTICA 2009: Proceedings of the
  9th Conference on Autonomous Robot Systems and Competitions}, P.~Gonçalves,
  P.~Torres, and C.~Alves, Eds.\hskip 1em plus 0.5em minus 0.4em\relax Castelo
  Branco, Portugal: Instituto Politécnico de Castelo Branco, 2009, pp. 59--65.

\bibitem{GarFraBru-etal2015techrep}
L.~Garattoni, G.~Francesca, A.~Brutschy, C.~Pinciroli, and M.~Birattari,
  ``Software infrastructure for e-puck (and {TAM}),'' IRIDIA, Université Libre
  de Bruxelles, Brussels, Belgium, Tech. Rep. TR/IRIDIA/2015-004, 2015.

\bibitem{PinTriOgr-etal2012SI}
C.~Pinciroli, V.~Trianni, R.~O'Grady, G.~Pini, A.~Brutschy, M.~Brambilla,
  N.~Mathews, E.~Ferrante, G.~A. Di~Caro, F.~Ducatelle, M.~Birattari, L.~M.
  Gambardella, and M.~Dorigo, ``{ARGoS}: a modular, parallel, multi-engine
  simulator for multi-robot systems,'' \emph{Swarm Intelligence}, vol.~6,
  no.~4, pp. 271--295, 2012.

\bibitem{TriLop2015PLOSONE}
V.~Trianni and M.~López-Ibáñez, ``Advantages of task-specific
  multi-objective optimisation in evolutionary robotics,'' \emph{PLOS ONE},
  vol.~10, no.~8, p. e0136406, 2015.

\bibitem{HasLigRudBir2021NATUCOM}
K.~Hasselmann, A.~Ligot, J.~Ruddick, and M.~Birattari, ``Empirical assessment
  and comparison of neuro-evolutionary methods for the automatic off-line
  design of robot swarms,'' \emph{Nature Communications}, vol.~12, p. 4345,
  2021.

\bibitem{LopDubPer-etal2016ORP}
M.~López-Ibáñez, J.~Dubois-Lacoste, L.~Pérez~Cáceres, M.~Birattari, and
  T.~Stützle, ``The irace package: iterated racing for automatic algorithm
  configuration,'' \emph{Operations Research Perspectives}, vol.~3, pp. 43--58,
  2016.

\bibitem{HanOst2001EC}
N.~Hansen and A.~Ostermeier, ``Completely derandomized self-adaptation in
  evolution strategies,'' \emph{Evolutionary Computation}, vol.~9, no.~2, pp.
  159--195, 2001.

\bibitem{GlaSchYi-etal2010gecco}
T.~Glasmachers, T.~Schaul, S.~Yi, D.~Wierstra, and J.~Schmidhuber,
  ``Exponential natural evolution strategies,'' in \emph{GECCO'10: Proceedings
  of the 12th annual conference on Genetic and evolutionary computation}.\hskip
  1em plus 0.5em minus 0.4em\relax New York, NY, USA: ACM, 2010, pp. 393--400.

\bibitem{StaMii2002EC}
K.~O. Stanley and R.~Miikkulainen, ``Evolving neural networks through
  augmenting topologies,'' \emph{Evolutionary Computation}, vol.~10, no.~2, pp.
  99--127, 2002.

\bibitem{FraBraBru-etal2014ants}
G.~Francesca, M.~Brambilla, A.~Brutschy, L.~Garattoni, R.~Miletitch,
  G.~Podevijn, A.~Reina, T.~Soleymani, M.~Salvaro, C.~Pinciroli, V.~Trianni,
  and M.~Birattari, ``An experiment in automatic design of robot swarms:
  {AutoMoDe-Vanilla}, {EvoStick}, and human experts,'' in \emph{Swarm
  Intelligence: 9th International Conference, ANTS 2014}, ser. Lecture Notes in
  Computer Science, M.~Dorigo, M.~Birattari, S.~Garnier, H.~Hamann,
  M.~Montes~de Oca, C.~Solnon, and T.~Stützle, Eds., vol. 8667.\hskip 1em plus
  0.5em minus 0.4em\relax Cham, Switzerland: Springer International Publishing,
  2014, pp. 25--37.

\bibitem{KegGarBir2019taros}
M.~Kegeleirs, D.~Garzón~Ramos, and M.~Birattari, ``Random walk exploration for
  swarm mapping,'' in \emph{Towards Autonomous Robotic Systems: 20th Annual
  Conference, TAROS 2019}, ser. Lecture Notes in Computer Science,
  K.~Althoefer, J.~Konstantinova, and K.~Zhang, Eds., vol. 11650.\hskip 1em
  plus 0.5em minus 0.4em\relax Cham, Switzerland: Springer, 2019, pp. 211--222.

\bibitem{Con1999book}
W.~J. Conover, \emph{Practical Nonparametric Statistics}, 3rd~ed., ser. Wiley
  Series in Probability and Statistics.\hskip 1em plus 0.5em minus 0.4em\relax
  New York, NY, USA: John Wiley \& Sons, 1999.

\bibitem{LigBir2020SI}
A.~Ligot and M.~Birattari, ``Simulation-only experiments to mimic the effects
  of the reality gap in the automatic design of robot swarms,'' \emph{Swarm
  Intelligence}, vol.~14, pp. 1--24, 2020.

\bibitem{StrCasDor2020FRAI}
V.~Strobel, E.~Castelló~Ferrer, and M.~Dorigo, ``Blockchain technology secures
  robot swarms: a comparison of consensus protocols and their resilience to
  byzantine robots,'' \emph{Frontiers in Robotics and AI}, vol.~7, p.~54, 2020.

\bibitem{CasHarPenDor2021SCIROB}
E.~Castelló~Ferrer, T.~Hardjono, A.~Pentland, and M.~Dorigo, ``Secure and
  secret cooperation in robot swarms,'' \emph{Science Robotics}, vol.~6,
  no.~56, p. eabf1538, 2023.

\bibitem{StrPacDor2023SCIROB}
V.~Strobel, A.~Pachecho, and M.~Dorigo, ``Robot swarms neutralize harmful
  byzantine robots using a blockchain-based token economy,'' \emph{Science
  Robotics}, vol.~8, no.~79, p. eabm4636, 2023.

\bibitem{ReiZakDemFer2023COMPHY}
A.~Reina, R.~Zakir, G.~De~Masi, and E.~Ferrante, ``Cross-inhibition leads to
  group consensus despite the presence of strongly opinionated minorities and
  asocial behaviour,'' \emph{Communications Physics}, vol.~6, p. 236, 2023.

\end{thebibliography}

\end{document}